\documentclass{article} 
\usepackage[T1]{fontenc}
\usepackage{iclr2027_conference,times}
\font\iclrtenhv=cmr10 at 8pt

\usepackage{amsmath,amsfonts,bm}

\newcommand{\figleft}{{\em (Left)}}

\newcommand{\figright}{{\em (Right)}}

\def\eqref#1{equation~\ref{#1}}

\def\1{\bm{1}}

\DeclareMathAlphabet{\mathsfit}{\encodingdefault}{\sfdefault}{m}{sl}
\SetMathAlphabet{\mathsfit}{bold}{\encodingdefault}{\sfdefault}{bx}{n}

\newcommand{\E}{\mathbb{E}}

\newcommand{\R}{\mathbb{R}}

\newcommand{\Var}{\mathrm{Var}}

\usepackage{placeins}
\usepackage{graphicx}
\usepackage{booktabs}
\usepackage{longtable}
\usepackage{amsthm}
\usepackage{algorithm}
\usepackage{algpseudocode}
\usepackage{hyperref}
\hypersetup{hidelinks,pdftitle={Recursive Self-Improvement on a Budget: Scoring Code Edits Before They Run}}
\usepackage{url}
\usepackage{cleveref}
\algrenewcommand\algorithmicrequire{\textbf{Input:}}
\algrenewcommand\algorithmicensure{\textbf{Output:}}

\theoremstyle{plain}
\newtheorem{theorem}{Theorem}

\theoremstyle{definition}
\newtheorem{definition}{Definition}
\newtheorem{assumption}{Assumption}

\crefname{theorem}{theorem}{theorems}
\crefname{lemma}{lemma}{lemmas}
\crefname{proposition}{proposition}{propositions}
\crefname{definition}{definition}{definitions}
\crefname{assumption}{assumption}{assumptions}
\crefname{equation}{equation}{equations}
\crefname{figure}{figure}{figures}
\crefname{table}{table}{tables}
\crefname{section}{section}{sections}
\crefname{appendix}{appendix}{appendices}
\crefname{algorithm}{algorithm}{algorithms}

\title{Recursive Self-Improvement on a Budget: Scoring Code Edits Before They Run}

\author{Erfan Darzi$^{1,2}$, \quad Sree Bhargavi Balija$^{3}$, \quad Sheng He$^{1}$ \\[4pt]
$^{1}$Harvard University \quad $^{2}$Massachusetts Institute of Technology \quad $^{3}$UC San Diego
}

\iclrfinalcopy 
\begin{document}

\maketitle
\pdfbookmark[0]{Recursive Self-Improvement on a Budget: Scoring Code Edits Before They Run}{bm:title}

\begin{abstract}
A coding agent that rewrites its own scaffold has to choose which proposed edit to expand before that edit has been evaluated, and the search then has to return one agent.
Each run fits one linear readout on the parent's steering-split solve rate and eight coordinates of the proposed diff, from 96 scaffold edits labeled by four evaluations of a 16-instance fit split, and then freezes the coefficients.
HumanEval+ reserves 48 instances as the steering split and Polyglot reserves 64.
That split scores parents and keeps an archive of four nodes.
Huxley--G\"odel Machine selection draws a parent by Thompson sampling on the steering counts of nodes that have already been evaluated, and it expands the child that parent proposes.
Every method then returns the earliest evaluated node whose steering-split solve count is maximal.
The tables report the held-out score of that node on the other 100 HumanEval+ instances and the other 145 Polyglot instances, next to the root scaffold and the calibration edit with the highest fit-split label.
Twelve runs of GPT-4o return a node at 87.0\% on held-out HumanEval+, against 84.2\% for the parent-rate ablation.
On that split Huxley--G\"odel Machine selection returns 85.1\%, the root is at 81.2\%, and the calibration-best edit is at 82.8\%.
On the same split Claude Sonnet 4.5 returns 86.6\%, beside 86.5\% for Huxley--G\"odel Machine selection, 85.3\% for the parent-rate ablation, and 85.1\% for the calibration-best edit.
\end{abstract}

\section{Introduction}
\label{sec:intro}

\begin{figure}[t]
\centering
\includegraphics[width=0.92\linewidth]{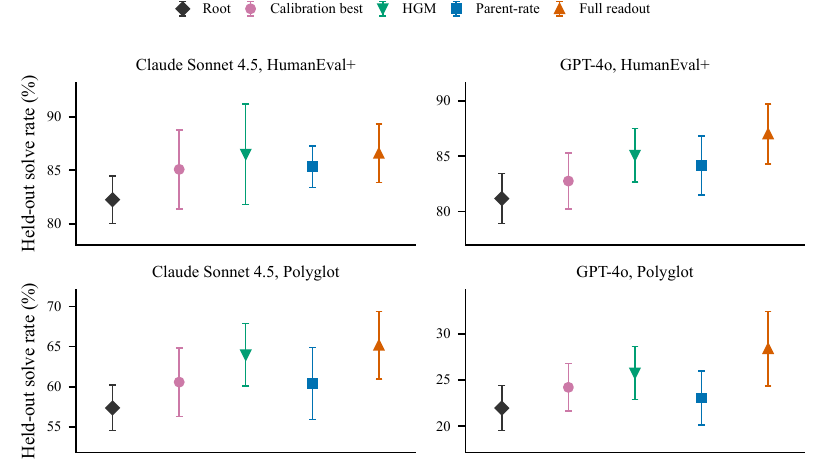}
\caption{Held-out solve rate of the returned node, twelve runs.
Points are means and bars are run standard deviations.
For the readout, the parent-rate ablation, and Huxley--G\"odel Machine selection, the node is the earliest evaluated node whose steering-split solve count is maximal.
The root is the starting scaffold, and calibration best is the edit with the highest fit-split label.
On HumanEval+ with GPT-4o the readout is the highest mean.
On HumanEval+ with Claude Sonnet 4.5 the readout sits with Huxley--G\"odel Machine selection, the parent-rate node, and the calibration-best edit.
\Cref{tab:main} reports these means.}
\label{fig:teaser}
\end{figure}

A coding agent that rewrites its own scaffold has to rank the edits it proposes before any of them has been evaluated, because the checker runs on one child per keep step and the child that is kept becomes a parent of the edits that follow.
The Darwin G\"odel Machine adds a child to its archive only after that child has been evaluated, and it samples parents from observed scores \citep{zhang2026darwin}.
The Huxley--G\"odel Machine draws a parent by Thompson sampling on the success and failure counts of evaluated nodes and expands the child that parent proposes \citep{wang2026huxley}.
AIDE scores a candidate after the evaluator has run it \citep{jiang2025aide}.
Surrogate-assisted evolution fits a predictor of fitness from earlier evaluations of similar individuals \citep{jin2011surrogate}.
The readout is that kind of predictor, applied to the diff before the child is expanded.
It is a linear score on the parent's steering-split solve rate and eight coordinates of the diff.
The ablation uses the same archive and expands by the parent's steering rate alone.
When the keep steps end, the readout, the parent-rate ablation, and Huxley--G\"odel Machine selection each return the earliest evaluated node whose steering-split solve count is maximal.
The held-out score of that node is the comparison, together with the root and the calibration edit whose fit-split label is highest.
Huxley--G\"odel Machine selection draws $\theta$ from $\mathrm{Beta}(1+n_s, 1+n_f)$ for each evaluated node, using that node's steering counts, and expands the child of the parent with the largest draw.
Every run of every method spends the same number of checker evaluations: the readout and the ablation label 96 calibration edits and then expand 24 children, and Huxley--G\"odel Machine selection expands 144 children on Polyglot and 176 on HumanEval+.
Claude Sonnet 4.5 and GPT-4o (\texttt{gpt-4o-2024-08-06}) are fit separately \citep{openai2024gpt4o}.
The held-out splits are 100 HumanEval+ instances and 145 Polyglot instances \citep{liu2023evalplus,gauthier2024polyglot}.
\Cref{tab:main} is the held-out score of the returned node for the readout, the parent-rate ablation, and Huxley--G\"odel Machine selection.
\Cref{tab:birth} is the child the readout and the parent-rate ablation expand at the first keep step, and \Cref{fig:teaser} plots the returned node.
\Cref{thm:birth} gives the mean of that parent draw, $(1+n_s)/(2+n_s+n_f)$, on the steering counts of an evaluated node.
The proof is in Appendix~\ref{app:proofs}.

\begin{itemize}
\item The readout ranks eight scaffold diffs at each keep step, and every method returns the earliest evaluated node whose steering-split solve count is maximal (\Cref{alg:search}, \Cref{fig:overview,fig:compare}).
\item Huxley--G\"odel Machine selection draws a parent from $\mathrm{Beta}(1+n_s, 1+n_f)$ on the steering counts of evaluated nodes and expands that parent's child (\Cref{thm:birth}, \Cref{fig:compare}).
\item On held-out HumanEval+, GPT-4o's returned node leads the parent-rate ablation, and the same table reports Huxley--G\"odel Machine selection, the root, and the calibration-best edit (\Cref{tab:main}, \Cref{fig:teaser}).
\end{itemize}

\section{Method}
\label{sec:method}

\subsection{Birth-time readout}
\label{sec:clade}

The root is a scaffold: a system prompt and a loop that writes a solution, runs the unit tests, and retries once.
A proposal is a unified diff of that scaffold.
The readout scores that diff from the parent's steering-split solve rate and from eight coordinates of the diff, before the diff is expanded.
The diff is embedded by \texttt{text-embedding-3-small} as a vector in $\R^{1536}$.
Each run draws its own projection, with i.i.d.\ standard normal entries and rows scaled to unit length, and it maps that vector onto the coordinates $e$.
The birth feature of the full readout is $X=(R_{\mathrm{par}}, e)\in\R^{9}$.
Both $R_{\mathrm{par}}$ and the label $R$ are solve rates in $[0,1]$.

Before the search, each run calibrates its own readout.
It proposes 96 scaffold edits, 48 of the root and two from each of 24 first-generation children of the root, and labels each edited child by its mean solve rate over four evaluations of a fixed 16-instance split of the benchmark being run.
Each of those 24 parents is evaluated once on the steering split, so $R_{\mathrm{par}}$ is measured for every calibration parent.
Because those parents do not share one rate, the least-squares slope on $R_{\mathrm{par}}$ is identified, and on these pairs that slope is positive.
Each generator is fit separately on each benchmark.
The full readout minimizes $\sum_{i=1}^{96}(R_i - w^{\top} X_i - b)^2$ and scores a proposal by $\widehat{v}(X)=w^{\top} X + b$.
The ablation minimizes the same squared error on $R_{\mathrm{par}}$ alone,
\begin{equation}
\sum_{i=1}^{96}\bigl(R_i - w_r R_{\mathrm{par},i} - b_r\bigr)^2,
\label{eq:ablation}
\end{equation}
and scores a proposal by $\widehat{v}_r(R_{\mathrm{par}})=w_r R_{\mathrm{par}}+b_r$.
The coefficients stay fixed for the 24 keep steps.

During the search, $R_{\mathrm{par}}$ is the parent's solve rate on the steering split, 48 HumanEval+ instances or 64 Polyglot instances, disjoint from the 16 fit instances.
The readout and the ablation keep the four evaluated nodes with the largest solve rate on that steering split.
Huxley--G\"odel Machine selection keeps the tree.
For each evaluated node it draws $\theta$ from $\mathrm{Beta}(1+n_s, 1+n_f)$ on the success and failure counts of that node's clade on the steering split, and it expands the child proposed by the parent with the largest draw.
The steering split chooses the parents, the archive, and the returned node.
The fit, the archive, and the parent draw read the steering split.
The tables report the held-out score of the node that rule returns.

\begin{figure}[t]
\centering
\includegraphics[width=\linewidth]{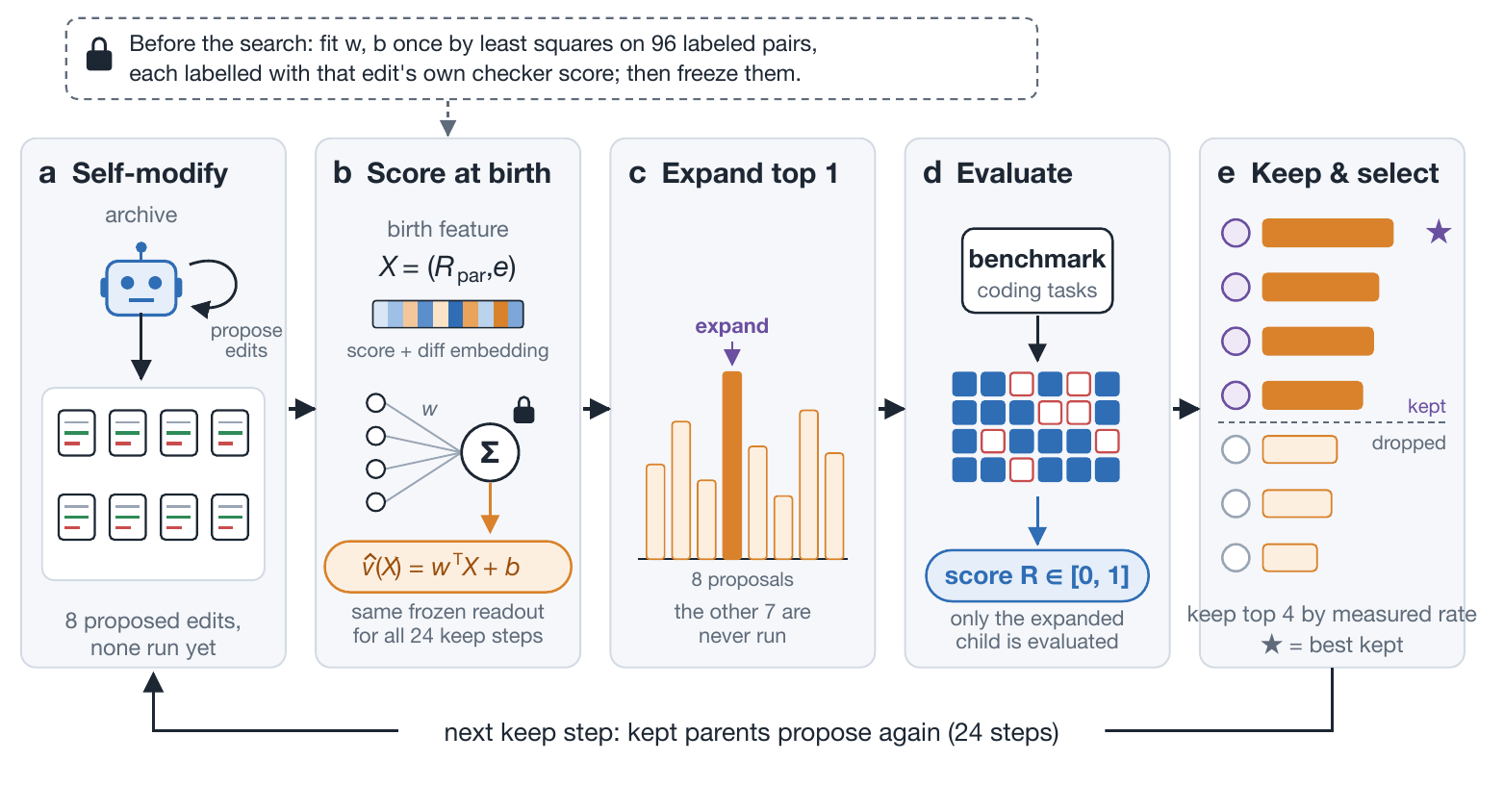}
\caption{One keep step of the full readout.
The eight proposals are divided across the archive.
The birth feature is $X=(R_{\mathrm{par}}, e)$, with $e$ the eight coordinates from the run's projection.
The readout expands the highest score.
The archive keeps four nodes by the steering-split solve rate.}
\label{fig:overview}
\end{figure}

Each keep step of the readout and of the ablation divides eight proposals as evenly as possible across the archive and expands the argmax of that method's score.
On a tie the earliest proposal is expanded.
Proposals are decoded at temperature $1.0$, so the eight diffs of one keep step differ in what they change and by how much, and a diff that does not apply cleanly is discarded and redrawn.
The archive begins as the evaluated root, so the first eight proposals share $R_{\mathrm{par}}$ and the ablation's scores tie.
The weights on $e$ are then what ranks those eight diffs.
From the next keep step the archive holds more than one parent.
The positive slope $w_r$ orders the ablation's proposals by the parent's steering-split rate, and the weights on $e$ can prefer a diff from a lower-scoring parent.
The readout and the ablation run for $H=24$ keep steps.
Every method evaluates one child on the steering split at each keep step.
When those steps end, every method returns the earliest evaluated node whose steering-split solve count is maximal, and the root remains eligible.
\Cref{fig:overview} is one keep step of the full readout.

\subsection{Search loop}
\label{sec:loop}

\begin{figure}[t]
\centering
\includegraphics[width=\linewidth]{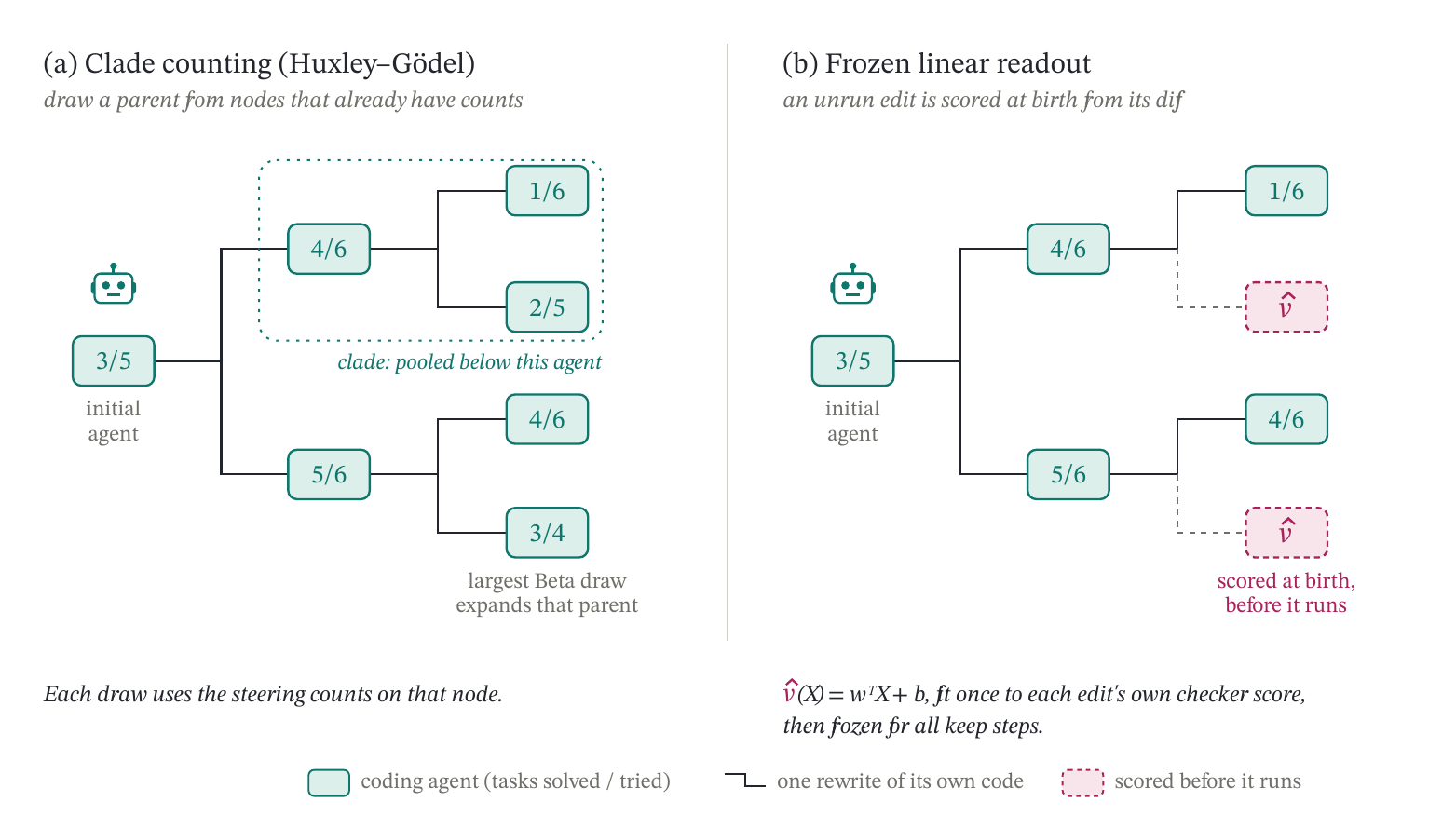}
\caption{Two ways to choose the next parent.
\figleft{} Huxley--G\"odel Machine selection draws $\theta$ for each evaluated node from its steering counts and expands the child of the largest draw.
\figright{} The full readout scores the edit from $X=(R_{\mathrm{par}}, e)$ by $\widehat{v}(X) = w^{\top} X + b$.}
\label{fig:compare}
\end{figure}

\Cref{alg:search} is the search, and the evaluator on each benchmark is that benchmark's checker.
\Cref{fig:compare} places Huxley--G\"odel Machine selection beside the full readout.
The parent draw uses the steering counts of evaluated nodes, and \Cref{thm:birth} gives its mean.

\begin{algorithm}[t]
\caption{Birth-time readout search. The loop stops after $H$ keep steps, with $H$ set so that every method spends the same number of checker evaluations per run: $H=24$ for the readout and the ablation, and $H=144$ on Polyglot or $176$ on HumanEval+ for Huxley--G\"odel Machine selection.}
\label{alg:search}
\begin{algorithmic}[1]
\Require \parbox[t]{0.92\linewidth}{\raggedright Generator $G$, Claude Sonnet 4.5 (\texttt{claude-sonnet-4-5-20250929}) or GPT-4o (\texttt{gpt-4o-2024-08-06}); checker $C$; horizon $H$; eight proposals at decoding temperature $1.0$; archive width $4$. Edits are unified diffs of that scaffold. Each run draws its own projection mapping the embedding onto $e\in\R^{8}$. The full readout uses $X=(R_{\mathrm{par}}, e)$. The steering split is 48 HumanEval+ instances or 64 Polyglot instances.\strut}
\Ensure The evaluated node with the largest steering-split solve count, scored on the held-out split.
\State On this generator and this benchmark, fit $(w, b)$ on $X$ and $(w_r, b_r)$ on $R_{\mathrm{par}}$ alone, using the 96 pairs of \Cref{sec:clade}. Hold both fits fixed.
\State Evaluate the root with $C$ on the steering split, set the archive to that root, and set its clade counts from that evaluation.
\While{fewer than $H$ keep steps have completed}
  \If{the method is the full readout or the parent-rate ablation}
    \State Propose eight edits, divided as evenly as possible across the archive. Redraw a diff that does not apply.
    \State Score each edit by $\widehat{v}(X)$ or by $\widehat{v}_r(R_{\mathrm{par}})$, according to the method.
    \State Expand the edit with the largest score. On a tie, expand the earliest proposal.
    \State Evaluate that child with $C$ on the steering split.
    \State Keep the four evaluated nodes with the largest steering-split solve rate.
  \Else
    \State Draw $\theta\sim\mathrm{Beta}(1+n_s,1+n_f)$ for each evaluated node and expand the child of the parent with the largest draw.
    \State Evaluate that child with $C$ on the steering split and add the outcomes to the clade counts of its ancestors.
  \EndIf
\EndWhile
\State Return the evaluated node with the largest steering-split solve count. On a tie, return the earliest node.
\State Score that node on the held-out split.
\end{algorithmic}
\end{algorithm}

\subsection{Birth-time law}
\label{sec:birthlaw}

Huxley--G\"odel Machine selection uses the case $\tau = 1$ of the law below.
The density, the ratio, and the event of zero clade counts are the objects the theorem uses.

\begin{definition}[Beta draw]
\label{def:beta}
For $\tau > 0$ and integers $n_s, n_f \geq 0$, the law of $\theta_{n_s,n_f}^{(\tau)}$ is the probability measure on $(0,1)$ with density
$B(\alpha,\beta)^{-1} u^{\alpha-1}(1-u)^{\beta-1}$,
where $\alpha = \tau(1+n_s)$, $\beta = \tau(1+n_f)$, and $B(\alpha,\beta) = \int_0^1 u^{\alpha-1}(1-u)^{\beta-1}\, du \in (0,\infty)$.
\end{definition}

\begin{definition}[Ratio estimator]
If $n_s + n_f \geq 1$, set $\rho(n_s,n_f) = n_s/(n_s+n_f)$.
The expression is not a real number when $n_s = n_f = 0$.
\end{definition}

\begin{definition}[Birth time]
\label{def:birth}
A birth time of a node is an event on which that node's clade counts, in the sense of the subtree rooted at the node, including the node, satisfy $n_s = n_f = 0$.
The code-change feature may still vary.
\end{definition}

\begin{assumption}[Positive scheduler]
\label{ass:tau}
$\tau > 0$.
\end{assumption}

\begin{assumption}[Birth-time zero counts]
\label{ass:birth}
The node under discussion is at a birth time in the sense of \Cref{def:birth}.
\end{assumption}

\begin{theorem}
\label{thm:birth}
Assume \Cref{ass:tau,ass:birth}.
On the event of \Cref{ass:birth}, $\theta_{0,0}^{(\tau)}$ has law $\mathrm{Beta}(\tau, \tau)$.
This law does not depend on $X$.
Its mean is $m(0,0) = 1/2$, and its variance is $1/(4(2\tau+1))$.
For all integers $n_s, n_f \geq 0$ and every $\tau > 0$,
$\E[\theta_{n_s,n_f}^{(\tau)}] = m(n_s,n_f) = (1+n_s)/(2+n_s+n_f)$.
The ratio $\rho(0,0)$ is undefined.
The case $\tau = 1$ is a $\mathrm{Beta}(1,1)$ draw, uniform on $(0,1)$, with mean $1/2$ and variance $1/12$.
\end{theorem}

The experiments use $\tau = 1$.
Huxley--G\"odel Machine selection draws $\theta$ from this law for each evaluated node, from the success and failure counts on that node's steering evaluation, and expands the child of the parent with the largest draw.
The mean of the draw is $m(n_s,n_f)=(1+n_s)/(2+n_s+n_f)$.
The proof is in Appendix~\ref{app:proofs}.

\section{Experiments}
\label{sec:exp}
\FloatBarrier

The held-out splits are the 100 HumanEval+ instances and the 145 Polyglot instances that remain after the 16-instance fit split and the steering split \citep{liu2023evalplus,chen2021codex,gauthier2024polyglot}.
The steering splits contain 48 and 64 instances.
Claude Sonnet 4.5 (\texttt{claude-sonnet-4-5-20250929}) and GPT-4o (\texttt{gpt-4o-2024-08-06}) are fit separately and run separately.
Each cell is twelve runs.
Each run draws its own projection, calibrates its own coefficients on its own 96 edits, and draws its own proposals.
Proposals are decoded at temperature $1.0$ and the scaffold's solutions at $0.2$.
In four-fold cross-validation on each run's calibration edits, the fitted score ranks the measured labels at Spearman $\rho$ between $0.18$ and $0.34$ across the four settings.
Every run of every method spends the same number of checker evaluations: 9{,}280 on Polyglot and 8{,}496 on HumanEval+.
A run of the readout or of the ablation spends 6{,}144 of them labeling the calibration edits, 1{,}536 on Polyglot or 1{,}152 on HumanEval+ evaluating the 24 calibration parents on the steering split, 64 or 48 evaluating the root, and the rest on its 24 keep steps.
Huxley--G\"odel Machine selection evaluates the root and expands 144 children on Polyglot and 176 on HumanEval+, reaching the same totals.
The calibration stage is itself a search: the edit with the highest measured label is a candidate agent, and \Cref{tab:main} reports it as the calibration-best baseline.
After the keep steps, the readout, the parent-rate ablation, and Huxley--G\"odel Machine selection each return one node, the earliest evaluated node whose steering-split solve count is maximal.
\Cref{tab:main} is the held-out score of that node, of the root, and of the calibration-best edit.
\Cref{tab:birth} is the held-out score of the child the readout and the parent-rate ablation expand at step 1.
Printed means and run standard deviations follow the aggregation protocol in Appendix~\ref{app:details}.
A paired interval is the 95\% Student interval on the twelve run-wise differences, with eleven degrees of freedom.
The one-sided permutation probability is the fraction of the 4096 sign flips whose signed sum is at least the observed sum.

\begin{table}[t]
\caption{Held-out solve rate of the returned node.
For the readout, the parent-rate ablation, and Huxley--G\"odel Machine selection, the node is the earliest evaluated node whose steering-split solve count is maximal.
The root row is the starting scaffold, and calibration best is the edit with the highest measured label among the run's 96 calibration edits.
The standard deviation is across the twelve runs.
Bold marks the highest mean and underline marks the second highest.}
\label{tab:main}
\centering
\footnotesize
\setlength{\tabcolsep}{3.5pt}
\begin{tabular}{lcccc}
\toprule
& \multicolumn{2}{c}{Claude Sonnet 4.5} & \multicolumn{2}{c}{GPT-4o} \\
\cmidrule(lr){2-3} \cmidrule(lr){4-5}
Method & HumanEval+ $\uparrow$ & Polyglot $\uparrow$ & HumanEval+ $\uparrow$ & Polyglot $\uparrow$ \\
\midrule
Root & $82.3 \pm 2.22$ & $57.4 \pm 2.83$ & $81.2 \pm 2.25$ & $22.0 \pm 2.46$ \\
Calibration best & $85.1 \pm 3.68$ & $60.6 \pm 4.26$ & $82.8 \pm 2.53$ & $24.2 \pm 2.57$ \\
HGM & $\underline{86.5 \pm 4.68}$ & $\underline{64.0 \pm 3.88}$ & $\underline{85.1 \pm 2.43}$ & $\underline{25.7 \pm 2.89}$ \\
$R_{\mathrm{par}}$-only & $85.3 \pm 1.92$ & $60.4 \pm 4.46$ & $84.2 \pm 2.66$ & $23.0 \pm 2.93$ \\
Full readout & $\mathbf{86.6 \pm 2.75}$ & $\mathbf{65.2 \pm 4.22}$ & $\mathbf{87.0 \pm 2.70}$ & $\mathbf{28.4 \pm 4.04}$ \\
\bottomrule
\end{tabular}
\end{table}

\begin{table}[t]
\caption{Held-out solve rate of the child expanded at keep step 1, when the archive holds only the root.
Entries are the mean $\pm$ sample standard deviation over twelve runs.
Bold marks the highest mean.}
\label{tab:birth}
\centering
\footnotesize
\setlength{\tabcolsep}{4pt}
\begin{tabular}{lcccc}
\toprule
& \multicolumn{2}{c}{Claude Sonnet 4.5} & \multicolumn{2}{c}{GPT-4o} \\
\cmidrule(lr){2-3} \cmidrule(lr){4-5}
Method & HumanEval+ $\uparrow$ & Polyglot $\uparrow$ & HumanEval+ $\uparrow$ & Polyglot $\uparrow$ \\
\midrule
Full readout & $\mathbf{83.8 \pm 2.09}$ & $\mathbf{60.1 \pm 4.15}$ & $\mathbf{82.3 \pm 3.02}$ & $\mathbf{25.2 \pm 2.17}$ \\
$R_{\mathrm{par}}$-only & $81.7 \pm 3.37$ & $57.4 \pm 4.00$ & $80.6 \pm 3.75$ & $22.2 \pm 3.72$ \\
\bottomrule
\end{tabular}
\end{table}

At keep step 1 the archive holds only the root, so the eight proposals share one parent rate.
The readout ranks them by the coordinates of the diff.
The parent-rate scores tie, and the ablation expands the earliest proposal.
On held-out Polyglot with Claude Sonnet 4.5 those children score 60.1\% and 57.4\% (\Cref{tab:birth}).

\Cref{tab:main} is the held-out score of the node returned after the keep steps.
On HumanEval+ with GPT-4o that node scores 87.0\%, the parent-rate node 84.2\%, Huxley--G\"odel Machine selection 85.1\%, the calibration-best edit 82.8\%, and the root 81.2\%.
The paired difference against the parent-rate node is 2.83 points, with interval $[0.65, 5.01]$ and permutation probability $42/4096$.
The paired difference against Huxley--G\"odel Machine selection is 1.92 points, with interval $[-0.87, 4.70]$ and permutation probability $368/4096$.
On the same split Claude Sonnet 4.5 returns 86.6\%, Huxley--G\"odel Machine selection 86.5\%, the parent-rate node 85.3\%, and the calibration-best edit 85.1\%.
The paired difference against the parent-rate node is 1.25 points, interval $[-0.83, 3.33]$, probability $490/4096$, and against the calibration-best edit it is 1.50 points, interval $[-1.68, 4.68]$, probability $697/4096$.
These comparisons are paired within runs.
On Polyglot, Claude Sonnet 4.5 returns 65.2\% for the readout, 64.0\% for Huxley--G\"odel Machine selection, 60.4\% for parent-rate expansion, 60.6\% for the calibration-best edit, and 57.4\% for the root.
GPT-4o returns 28.4\%, 25.7\%, 23.0\%, 24.2\%, and 22.0\% on that split.
\Cref{fig:teaser} plots the returned node.
\FloatBarrier

\section{Related work}
\label{sec:related}

G\"odel machines accept a self-rewrite only after a proof of higher expected utility \citep{schmidhuber2007godel}.
\citet{good1966speculations} and \citet{clune2019aigas} discuss ultraintelligent machines and AI-generating algorithms.
The Self-Taught Optimizer (STOP), Automated Design of Agentic Systems (ADAS), and the Self-Improving Coding Agent (SICA) edit a scaffold or a codebase; SICA reports 17\% to 53\% on a random SWE-bench Verified subset \citep{zelikman2023stop,hu2024automated,robeyns2025self}.
The G\"odel Agent searches a space of agent designs \citep{yin2025godel}.
With Claude 3.5 Sonnet (new), DGM reports 20.0\% to 50.0\% on a staged SWE-bench subset \citep{zhang2026darwin}.
Wang et al.\ report 56.7\% on SWE-Verified-60 with a different generator \citep{wang2026huxley}.
In \Cref{tab:main} that parent draw runs for the full horizon, and the reported node is the earliest evaluated node whose steering-split solve count is maximal.
Liu et al.\ report 78.3\% on SWE-Verified-60 and 93.3\% on Polyglot with Qwen3.6-35B-A3B \citep{liu2026mendel}.
Lai et al.\ aggregate descendant outcomes into clade-level Beta beliefs and select nodes by Thompson sampling \citep{lai2026clade}.
AIDE scores a candidate only after an evaluator runs it, and AIDE\textsuperscript{2} keeps a self-rewrite only after a hidden evaluation \citep{jiang2025aide,srikanth2026recursive}.

\section{Conclusion}
\label{sec:conclusion}

The readout ranks scaffold diffs by the parent's steering-split solve rate and by eight coordinates of the diff, and every method returns the earliest evaluated node whose steering-split solve count is maximal.
On held-out HumanEval+ with GPT-4o, that node scores 87.0\%, against 84.2\% for the same rule after parent-rate expansion.
Huxley--G\"odel Machine selection returns 85.1\% on that split, the root is at 81.2\%, and the calibration-best edit is at 82.8\%.
On the same split Claude Sonnet 4.5 returns 86.6\%, beside 86.5\% for Huxley--G\"odel Machine selection, 85.3\% for parent-rate expansion, and 85.1\% for the calibration-best edit (\Cref{tab:main}, \Cref{fig:teaser}).

\section{Limitations}
\label{sec:limitations}

The readout is linear in eight coordinates of one projection per run, and the coefficients stay fixed through the 24 keep steps.
The fit uses 96 scaffold edits labeled on 16 instances, and the archive keeps four nodes.
The generators are Claude Sonnet 4.5 and GPT-4o (\texttt{gpt-4o-2024-08-06}), on HumanEval+ and Polyglot.
The steering split that keeps the archive is disjoint from the held-out split in \Cref{tab:main,tab:birth}.

\subsection*{Reproducibility statement}
The proof of \Cref{thm:birth} is in Appendix~\ref{app:proofs}. Evaluation and aggregation details for the twelve runs are in Appendix~\ref{app:details}. Appendix~\ref{app:prompts} gives the scaffold instruction.

\subsection*{Ethics statement}
The search can keep a lineage the checker later rejects.
On these benchmarks the immediate score is the automatic checker.

\bibliography{refs}
\bibliographystyle{iclr2027_conference}

\appendix
\section{Proofs}
\label{app:proofs}

\allowdisplaybreaks
The argument below is the proof of \Cref{thm:birth}.

\subsection{Proof of the birth-time law}

\emph{Step A, mean of a Beta law.}
Let $\alpha > 0$, $\beta > 0$, and write $I(\alpha,\beta) = \int_0^1 u^{\alpha-1}(1-u)^{\beta-1}\, du \in (0,\infty)$.
The mean of the corresponding Beta law is $I(\alpha+1,\beta)/I(\alpha,\beta)$.
Define $\phi(u) = u^{\alpha}(1-u)^{\beta}$ for $u \in [0,1]$, with $\phi(0) = \phi(1) = 0$ because $\alpha > 0$ and $\beta > 0$.
On $(0,1)$,
\begin{align*}
\phi'(u)
&= \alpha u^{\alpha-1}(1-u)^{\beta} - \beta u^{\alpha}(1-u)^{\beta-1} \\
&= \alpha\, u^{\alpha-1}(1-u)^{\beta-1} - (\alpha+\beta)\, u^{\alpha}(1-u)^{\beta-1}.
\end{align*}
For $0 < \delta < 1/2$, the fundamental theorem on $[\delta, 1-\delta]$ gives $\int_{\delta}^{1-\delta} \phi'(u)\, du = \phi(1-\delta) - \phi(\delta)$.
The right-hand side tends to $0$ as $\delta \downarrow 0$.
The two integrands on the expanded right-hand side are nonnegative, so the monotone convergence theorem yields $\alpha\, I(\alpha,\beta) - (\alpha+\beta)\, I(\alpha+1,\beta) = 0$, hence $I(\alpha+1,\beta)/I(\alpha,\beta) = \alpha/(\alpha+\beta)$.
\Cref{ass:tau} gives $\alpha = \tau(1+n_s) > 0$ and $\beta = \tau(1+n_f) > 0$, so
\[
\E\bigl[\theta_{n_s,n_f}^{(\tau)}\bigr]
= \frac{\tau(1+n_s)}{\tau(1+n_s)+\tau(1+n_f)}
= \frac{1+n_s}{2+n_s+n_f}.
\]
At $(n_s,n_f) = (0,0)$ the parameters are $(\tau,\tau)$ and the mean is $1/2$.
The same ratio at parameters $(\alpha+1,\beta)$ yields $\E[U^2] = \alpha(\alpha+1)/((\alpha+\beta)(\alpha+\beta+1))$.
Subtracting the squared mean produces
\[
\Var(U) = \frac{\alpha\beta}{(\alpha+\beta)^2(\alpha+\beta+1)}.
\]
At $\alpha = \beta = \tau$ this is $1/(4(2\tau+1))$.
For $\tau = 1$ the variance is $1/12$.
None of the parameters, the mean, or the variance reads $X$.
Under \Cref{ass:birth} the counts are $(0,0)$, so the sampling law is $\mathrm{Beta}(\tau,\tau)$ for every value of the code-change feature.
The symbol $\rho(0,0)$ would divide by zero and is left undefined.

\section{Experimental details}
\label{app:details}

Each cell of \Cref{tab:main,tab:birth} is twelve runs of Claude Sonnet 4.5 (\texttt{claude-sonnet-4-5-20250929}) or GPT-4o (\texttt{gpt-4o-2024-08-06}).
The steering split is 48 HumanEval+ instances or 64 Polyglot instances, and the held-out split is the other 100 or 145 instances.
Huxley--G\"odel Machine selection draws the parent from $\mathrm{Beta}(1+n_s, 1+n_f)$ on the steering counts of evaluated nodes.
The calibration-best row is the edit with the highest fit-split label, scored on the held-out split.

A printed accuracy is the one-decimal rounding of $100$ times the pooled solve count of the returned node divided by twelve times the split size, and the sample standard deviation is the ordinary standard deviation of the twelve run percentages.
For the readout and the parent-rate ablation the candidates are the root and the 24 expanded children.
For Huxley--G\"odel Machine selection the candidates are the root and the expanded children, 144 on Polyglot and 176 on HumanEval+.

\section{Prompts and reproductions}
\label{app:prompts}

Both generators are asked for a unified diff of the scaffold, at decoding temperature $1.0$; a diff that does not apply cleanly is discarded and redrawn.
The instruction is: ``The scaffold is a system prompt and a loop that writes a solution, runs the unit tests, and retries once. Return a unified diff of that scaffold.''

\end{document}